\documentclass[journal,twoside,web]{ieeecolor}
\usepackage{generic}
\usepackage{cite}
\usepackage{amsmath,amssymb,amsfonts}
\usepackage{graphicx}
\usepackage{textcomp}
\usepackage{makecell}
\usepackage{multirow}
\usepackage{pifont}
\usepackage{hyperref}
\usepackage{soul}
\soulregister\cite7 
\usepackage{xcolor}
\setul{0.5ex}{2pt}
\setstcolor{red}
\usepackage{soul}
\soulregister\ref7
\soulregister\pageref7
\usepackage[normalem]{ulem}

\def\BibTeX{{\rm B\kern-.05em{\sc i\kern-.025em b}\kern-.08em
    T\kern-.1667em\lower.7ex\hbox{E}\kern-.125emX}}
\begin{document}
\bstctlcite{bstctl:nodash}
\title{Towards Biosignals-Free Autonomous Prosthetic Hand Control via Imitation Learning}
\author{Kaijie Shi, Wanglong Lu, Hanli Zhao\MakeUppercase{*}, Vinicius Prado da Fonseca, \IEEEmembership{Member, IEEE}, Ting Zou, \IEEEmembership{Member, IEEE}, and Xianta Jiang\MakeUppercase{*}, \IEEEmembership{Senior Member, IEEE}
\thanks{\textcolor{red}{This article has been published in IEEE Transactions on Neural Systems and Rehabilitation Engineering.}}
\thanks{Submitted on March 19th, 2025. This work was supported in part by the Government of Canada's New Frontiers in Research Fund (NFRF, Grant No NFRFE-2022-00407) and Natural Sciences and Engineering Research Council of Canada's Research Tools and Instruments (NSERC RTI, Grant No  RTI-2022-00688).}
\thanks{This work involved human subjects or animals in its research. Approval of all ethical and experimental procedures and protocols was granted by the Memorial University Interdisciplinary Committee on Ethics in Human Research (20210316-SC).}
\thanks{
Kaijie Shi, Wanglong Lu are with Department of Computer Science, Memorial University of Newfoundland, St. John’s, NL A1B 3X5, Canada, and also with College of Computer Science and Artificial Intelligence, Wenzhou University, Wenzhou, 325000, China. (email: kaijies@mun.ca, wanglongl@mun.ca).
}
\thanks{
Hanli Zhao is with College of Computer Science and Artificial Intelligence, Wenzhou University, Wenzhou, 325000, China. (email: hanlizhao@wzu.edu.cn).
}
\thanks{
Vinicius Prado da Fonseca is with Department of Computer Science, Memorial University of Newfoundland, St. John’s, NL A1B 3X5, Canada. (email: vpradodafons@mun.ca).
}
\thanks{Ting Zou is with Department of Mechanical and Mechatronics Engineering, Memorial University of Newfoundland, St. John’s, NL A1B 3X5, Canada. (email: tzou@mun.ca). }
\thanks{
Xianta Jiang is with Department of Computer Science, Memorial University of Newfoundland, St. John’s, NL A1B 3X5, Canada. (email: xiantaj@mun.ca).
}
\thanks{*Corresponding authors: Hanli Zhao (email: hanlizhao@wzu.edu.cn); Xianta Jiang (email: xiantaj@mun.ca)}
}

\maketitle

\begin{abstract}
Limb loss affects millions globally, impairing physical function and reducing quality of life. 
Most traditional surface electromyographic (sEMG) and semi-autonomous methods require users to generate myoelectric signals for each control, imposing physically and mentally taxing demands.
This study aims to develop a fully autonomous control system that enables a prosthetic hand to automatically grasp and release objects of various shapes using only a camera attached to the wrist. 
By placing the hand near an object, the system will automatically execute grasping actions with a proper grip force in response to the hand’s movements and the environment. 
To release the object being grasped, just naturally place the object close to the table and the system will automatically open the hand. 
Such a system would provide individuals with limb loss with a very easy-to-use prosthetic control interface and may help reduce mental effort while using. To achieve this goal, we developed a teleoperation system to collect human demonstration data for training the prosthetic hand control model using imitation learning, which mimics the prosthetic hand actions from human. By training the model on data from a limited set of objects collected from a single participant's demonstration, we showed that the imitation learning algorithm can achieve high success rates and generalize effectively to new users and previously unseen objects with varying weights. The demonstrations are available at \url{https://sites.google.com/view/autonomous-prosthetic-hand}.

\end{abstract}


\section{Introduction}
\label{sec:introduction}
\IEEEPARstart {P}{rosthetic} hands play a crucial role for individuals who have lost their hands, helping regain the manipulation functionality and significantly improve their quality of life. However, controlling prosthetic hands remains a persistent challenge that limits their widespread acceptance. Traditional control methods, such as surface electromyography (sEMG) \cite{roche2014prosthetic, nasr2021musclenet, boostani2003evaluation, geethanjali2014low} and vision-based semi-autonomous\cite{shi2024semi, he2020vision, mcmullen2013demonstration, ng2024development} approaches, usually require users to repeatedly generate myoelectric signals for each operation, which could be physically and mentally demanding.

In particular, sEMG-based grasp pattern recognition for prosthetic hands has achieved impressive classification accuracy in lab settings~\cite{englehart2003robust}, and even home-settings experiments were conducted in \cite{hargrove2017myoelectric, simon2022user}, demonstrating that pattern recognition can outperform direct control. However, a few drawbacks still exist, such as noise and artifacts \cite{boyer2023reducing}, sensitivity to electrode placement \cite{keenan2011coherence, vieira2023sensitivity},
cross-talk issues \cite{mogk2003crosstalk, kong2010crosstalk}, individual physiological differences \cite{jarque2024does}, muscle fatigue \cite{jiang2012myoelectric, wang2021effect, fang2022simultaneous}, etc.
For example, different individuals or electrode locations \cite{jarque2024does} could cause input distribution shifts, and thus, sEMG classification models often fail when trained and tested on different individuals.
Meanwhile, users also experience muscle fatigue~\cite{jiang2012myoelectric} when required to generate specific sEMG signals repeatedly for each grasping action.

Recent advancements in vision-based semi-autonomous control methods offer significant potential in more natural prosthetic hand control~\cite{dovsen2010cognitive, he2020vision, zandigohar2024multimodal}. These systems normally integrate machine learning models to classify grip patterns based on images of target objects captured by a camera, in conjunction with sEMG signals from the user’s residual arm interpreting the user's intention to perform the gesture, such as closing, opening, and rotating the hand. 
Through systematic evaluation compared with manual control, semi-autonomous method can not only significantly reduce physical and mental workload \cite{starke2022semi}, but also require less time to accomplish the grasping tasks \cite{castro2022continuous}. Mouchoux et al. \cite{mouchoux2021artificial} also showed that semi-autonomous control with augmented reality feedback could enhance performance while reducing effort.
However, the system still relies on the user to generate distinct sEMG signals to convey the correct movement intention to make the system work.

In this study, we propose a novel autonomous prosthetic hand control system that can mimic natural human manipulation, the grasp, transfer, and release.
Unlike conventional systems that require selecting predefined grip patterns, our solution allows the prosthetic hand to autonomously adapt to objects of various shapes. More importantly, there is no need for explicit intervention from the user to open or close the hand during the reaching and grasping process. Instead, the user simply needs to bring the prosthetic hand near the target location, and the system will automatically execute grasping and releasing actions in response to the hand’s movements and the environment.

To achieve natural and autonomous prosthetic hand control, we employ imitation learning to enable the prosthetic system to learn directly from human demonstrations.
Specifically, we first employ a vision-based teleoperation system designed to collect comprehensive prosthetic hand control human demonstrations. This system records visual feedback obtained through wrist camera, proprioceptive information from the prosthetic hand, and prosthetic hand action gathered through teleoperation.
Next, we propose a generative visual-tactile-motor variational autoencoder (VTM-VAE) capable of reconstructing natural human actions conditioned upon visual and proprioceptive inputs. 
The system learns to dynamically adjust hand shape according to the target object, initiate and drive grasps based on the relative position between the hand and the target object, and apply appropriate gripping forces-behaviors acquired directly from demonstration data.
Moreover, leveraging this generative framework enables the model to effectively generalize from a limited set of training objects, to accurately predicting grasp behaviors for many previously unseen objects.

We conducted extensive experiments to evaluate the system.
The results show that the system achieves impressive results with human-like grasping and releasing on objects and generalizes well on unseen (untrained) objects, cluttered scenes, and grasping manipulations (e.g., handover tasks) without any model finetuning or biosignals engagement.
Furthermore, after training on data collected for a single participant, the model can generalize effectively to other participants.

\begin{figure*}[ht!]
    \centering
    \includegraphics[width=1.0\textwidth]{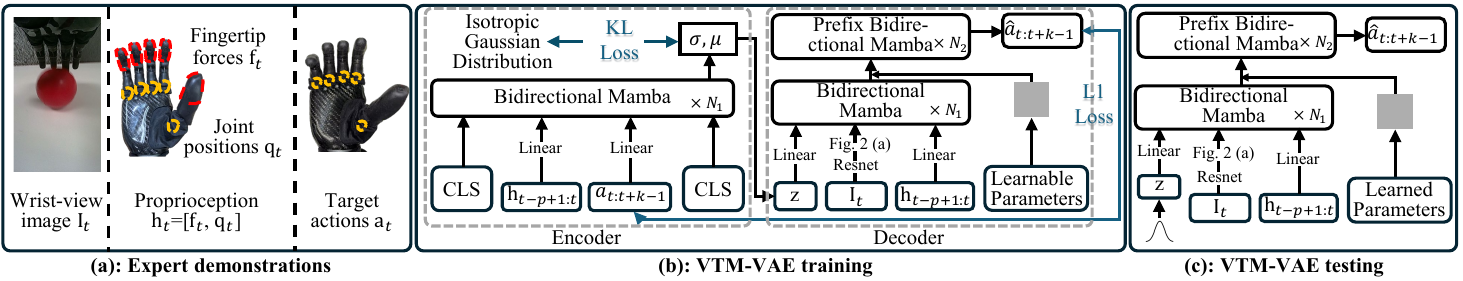}
    \caption{Overall pipeline of our method. The expert demonstrations (a) contains wrist-view image, proprioception, and target actions.
    Next, our visual-tactile-motor variational autoencoder (VTM-VAE) (b) learns to imitate the target actions with image, and proprioception as inputs. 
    In testing (c), the learned VTM-VAE decoder outputs smooth and accurate actions.
    }
    \label{fig:overall_pipeline}
\end{figure*}

To the best of our knowledge, this is the first fully autonomous prosthetic hand control system that uses only wrist-vision and proprioception to grasp and release objects, making it potentially applicable to real-world scenarios. This study presents the following key contributions:

\begin{enumerate}

\item \textbf{A biosignals-free autonomous prosthetic control method:} We introduce a novel approach that leverages vision and proprioception to control prosthetic hands, without any input of biosignals.
It provides an easy-to-use control method.

\item \textbf{Visual-Tactile-Motor Variational Autoencoder:} We propose a new generative visual-tactile-motor variational autoencoder designed for effective imitation learning, which enables the prosthetic hand to grasp and release objects naturally.

\item \textbf{Strong Generalization:} Our method demonstrates robust generalization across various objects and individuals, underscoring its potential for widespread real-world application.

\end{enumerate}

\section{Related work}

To control prosthetic hands, body-powered, myoelectric, and semi-autonomous methods have been widely studied. Body-powered control harnesses the user’s muscle power and body movements to actuate the prosthetic hand. Myoelectric approaches leverage muscle-generated signals to drive prosthetic movements, while semi-autonomous systems usually combine minimal muscle-generated signals or stump kinematics input with sensor data to enhance control. Despite significant progress in both areas, challenges remain in achieving natural, intuitive, and efficient operation in daily use.


Semi-autonomous control systems often use sEMG signals as simple triggers while relying on vision-based data to determine grasp patterns. A common approach uses computer vision to classify objects and then activates a pre-determined grasp pattern through sEMG signals \cite{shi2024semi}. In cluttered environments, object detection and eye tracking have been employed to select the target object \cite{he2020vision, mcmullen2013demonstration}. Furthermore, grasp affordance detection methods assign suitable grasp strategies to different object parts, enabling a single object to be associated with multiple grasp types \cite{ng2024development}.
In addition, to better integrate sEMG control with vision-based semi-autonoumous prosthetic hand control, Wang et al. \cite{wang2022phase, wang2022integrating} explored the best time for each module to take effect and the best fusion strategy to improve the performance.

Although these vision-based systems simplify the sEMG control problem to a binary task, either open or close the hand, users are still required to perform specific muscle activations during each grasping phase. The inherent challenges of sEMG-such as signal variability, noise, and muscle fatigue-remain unresolved, particularly when applied to a diverse range of users. Additionally, these systems typically rely on pre-defined gesture libraries, which restrict their ability to generalize to objects not included in those libraries.

More recently, Xu et al. \cite{xu2025powered} introduced a system that utilizes a head-mounted camera to achieve an autonomous grasping and lifting algorithm. However, due to the camera's fixed position on the user's head, the system requires users to maintain their gaze on the target object, thereby limiting natural movement. 
Similarly, although Sharif et al. \cite{sharif2020towards, sharif2021end} claimed to achieve autonomous control, their systems are confined to simulated environments \cite{sharif2020towards}, do not account for releasing tasks, and the reliance on a desk camera \cite{sharif2021end} further restricts the range of user activities.

Meanwhile, several notable works have exploited residual-limb kinematics to control a prosthetic hand.
Montagnani et al. \cite{montagnani2015exploiting} investigated the coordination between humeral orientation relative to the trunk and forearm pronation/supination angles during a broad spectrum of activities of daily living in healthy subjects, and found consistent postural synergies between these two segments across nearly all tasks examined.
Kuhn et al. \cite{kuhn2024synergy} and Merad et al. \cite{merad2020assessment} mapped stump movements into prosthetic joint actions via synergy models.
Bennett \& Goldfarb \cite{bennett2017imu} utilized a six-axis inertial measurement unit (IMU) to detect the upper-arm abduction/adduction, which was employed as the input to command the wrist rotational velocity.
Pilarski et al. \cite{pilarski2012dynamic} proposed a dynamic switching method that learns from real-time interactions to predict user behavior and improve the way users switch between functions of powered artificial limbs.

At the same time, proximity sensing and sensor fusion have enriched object perception capabilities. Mastinu et al. \cite{mastinu2024explorations} combined radar with low-resolution time-of-flight (ToF) imaging for real-time object recognition and auto-grasp planning. Heo et al. \cite{heo2023proximity} used embedded proximity sensors and point-cloud decision algorithms to infer intended grasp postures under a single-channel EMG interface. These approaches---stump kinematics, inertial measurement, proximity sensing, and mode-switch automation---together demonstrate a diverse ecosystem of semi-autonomous control paradigms that do not rely solely on cameras and computer vision.

In this study, we introduce a fully automatic control method for prosthetic hands that overcomes the limitations of sEMG-dependent systems. Our approach is designed to address the challenges of handling diverse objects by eliminating the need for pre-defined gesture libraries. The system supports essential tasks-such as grasping, lifting, and releasing-while employing a wrist-mounted camera design that enhances the user's range of motion and delivers a more natural, intuitive experience.

\section{Methodology}

The overall pipeline of our imitation learning algorithm is illustrated in Fig.\ref{fig:overall_pipeline}. The process begins with expert demonstration multi-modalities data (Fig.\ref{fig:overall_pipeline} (a)), which includes vision data from a wrist camera, tactile signals from the fingertips, motor information from the joints, and ground-truth actions provided by the expert.
To replicate expert prosthetic hand control, we introduce \textbf{visual-tactile-motor variational autoencoder (VTM-VAE)}, a generative learning framework, as depicted in Fig.~\ref{fig:overall_pipeline} (b). As a variant of the conditional variational autoencoder, the VTM-VAE learns the underlying distribution of human-demonstrated actions and proprioception during the training phase. During inference, it generates natural actions conditioned on wrist images and proprioceptive inputs (Fig.~\ref{fig:overall_pipeline} (c)).
To enable real-time inference and robust modeling, we define Central-aware Mamba (Fig.~\ref{fig:architecture}) to consist of Central-aware scan, Bidirectional Mamba, and Prefix Bidirectional Mamba, all of which are integrated into the VTM-VAE.

\subsection{Expert demonstrations}
Figure~\ref{fig:overall_pipeline} (a) illustrates an expert demonstration at a specific time step $t$. This demonstration comprises a wrist-view image $\mathbf{I}_t \in \mathbb{R}^{w \times h \times 3}$, joint positions $\mathbf{q}_t \in \mathbb{R}^{6}$, fingertip forces $\mathbf{f}_t \in \mathbb{R}^{30}$, and target actions $\mathbf{a}_t \in \mathbb{R}^{6}$.

\textbf{Wrist-view image $\mathbf{I}_t$.} In our approach, a camera mounted on the wrist captures images with a dimension of $320\times 240\times 3$. This configuration provides detailed visual information about both the prosthetic hand and its environment. Moreover, the wrist-mounted camera offers greater freedom and flexibility for the persons with limb loss compared to table-mounted or head-mounted setups.

\textbf{Joint positions $\mathbf{q}_t$.} The joint positions $\mathbf{q}_t$ indicate the state of the hand, such as whether it is open or closed. We follow the conventional approach in robot learning, where the joint positions represent the angles of motor rotation. In this study, the prosthetic hand (Ability hand, PSYONIC) features six degrees of freedom (DoFs): one for each finger, with the thumb having two DoFs.

\textbf{Fingertip forces $\mathbf{f}_t$.} Each fingertip is equipped with six built-in force-sensitive resistors (FSRs) that capture force information while manipulating, resulting in a total of 30-dimensional contact force measurements. This information directly indicates whether a finger is in contact with and how much force is applied to an object, a detail that is difficult to determine from occluded wrist-view images when grasping.

\textbf{Proprioception $\mathbf{h}_t$.} 
In our study, the joint positions $\mathbf{q}_t$ and the force measurements $\mathbf{f}_t$ are combined to form the proprioceptive data $\mathbf{h}_t$ at time step $t$.

\textbf{Target actions $\mathbf{a}_t$.} Target actions represent the desired joint positions that serve as commands for the prosthetic hand. The hand executes these commands using its integrated control system, and they are retargeted from human actions \cite{handa2020dexpilot} during the training demonstration. 
In our imitation learning framework, the algorithm learns these target actions to achieve a stable grasp.

\subsection{Visual-Tactile-Motor variational autoencoder} \label{sec:VTM-VAE}

During the training phase (Fig. \ref{fig:overall_pipeline} (b)), 
an encoder takes in target actions, tactile, and motor information, and converts them to a latent vector $\mathbf{z}$, which is restricted to close to standard isotropic Gaussian distribution. 
Then, a decoder takes in $\textbf{z}$, together with wrist image, tactile, and motor information, to reconstruct the corresponding target actions. 
The encoding and decoding processes can be expressed as:
\begin{equation}\label{equ:encoder}
	\begin{aligned}
        &(\log \boldsymbol{\sigma}^2, \boldsymbol{\mu}) = \textbf{E}(\operatorname{CLS}, \mathbf{h}_{t-p+1:t}, \mathbf{a}_{t:t+k-1}),\\
        &\hat{\mathbf{a}}_{t:t+k-1} = \textbf{D}(\mathbf{I}_t, \mathbf{h}_{t-p+1:t},\mathbf{z}),
        \mathbf{z} \sim \mathcal{N}(\boldsymbol{\mu},\boldsymbol{\sigma}^2), 
        \end{aligned}
\end{equation}
where $p$ represents past time-step, $k$ represents future time-step, $\operatorname{CLS}$ represents learnable parameters, $\textbf{E}$ and $\textbf{D}$ being the VTM-VAE encoder and decoder, respectively.
The specific details of the encoder and decoder's image feature processing (Central-aware scan) and backbone (Bidirectional Mamba and Prefix Bidirectional Mamba) are provided later in the Central-aware Mamba section.

The decoder's output target actions are expected to be as close as possible to the corresponding encoder's input target actions, facilitated by the optimization of a loss function:
\begin{align}
    L = \sum_{t=1}^{T} \left| \hat{\mathbf{a}}_{t:t+k-1} - \mathbf{a}_{t:t+k-1} \right| - \frac{1}{2} \left(1 + \log \boldsymbol{\sigma}^2 - \boldsymbol{\mu}^2 - \boldsymbol{\sigma}^2 \right),
    \label{eq: loss}
\end{align}
where $T$ is the total number of tuples ($\mathbf{I}_t$, $\mathbf{h}_t$, $\mathbf{a}_t$) in the collected dataset, $\mathbf{a}_{t:t+k-1}$ represents the target actions over $k$ consecutive frames, $\hat{\mathbf{a}}_{t:t+k-1}$ are the reconstructed actions, and the $\left| \cdot \right |$ denotes the L1 loss.

\subsubsection{VTM-VAE encoder} \label{sec: vtm-vae encoder}
As illustrated in Fig.~\ref{fig:overall_pipeline} (b), the input to the VTM-VAE encoder at timestep $t$ consists of two distinct $\operatorname{CLS}$ tokens, historical proprioception $\mathbf{h}_{t-p+1:t}$, and $k$ actions $\mathbf{a}_{t:t+k-1}$. Both $\mathbf{h}_{t-p+1:t}$ and $\mathbf{a}_{t:t+k-1}$ are projected to the same dimension $\mathbf{d}_{\text{model}}$ as the $\operatorname{CLS}$ tokens. These embeddings are then concatenated with the two $\operatorname{CLS}$ tokens. 
After positional encoding \cite{ke2020rethinking} is applied, the combined sequence is passed through the proposed Bidirectional Mamba module (Fig.~\ref{fig:architecture} (c)), repeated $N_1$ times, to produce an output. The first and last elements of this output are averaged along the first dimension and then projected to obtain $\boldsymbol{\mu}$ and $\log \boldsymbol{\sigma}^2$, which parameterize the prior isotropic Gaussian distribution.
Next, the reparameterization trick \cite{kingma2013auto} is applied to generate the latent variable $\mathbf{z}$, as defined by the following equation: 
\begin{align}
    \mathbf{z} = \boldsymbol{\mu} + \exp\left(\frac{1}{2} \log \boldsymbol{\sigma}^2\right) \cdot \boldsymbol{\epsilon} = \boldsymbol{\mu} + \boldsymbol{\sigma} \cdot \boldsymbol{\epsilon},
    \label{eq: reparametrize}
\end{align}
where $\boldsymbol{\epsilon} \sim \mathcal{N}(\mathbf{0}, \mathbf{I})$.

\subsubsection{VTM-VAE decoder} \label{sec: vtm-vae decoder}
\begin{figure}[t!]
    \centering  \includegraphics[width=0.4\textwidth]{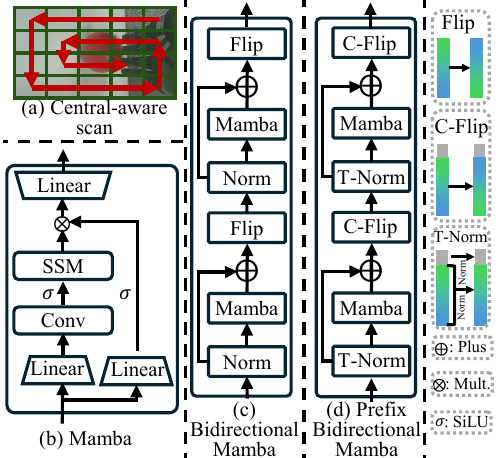}
    \caption{An illustration of the components of Central-aware Mamba. We proposed central-aware scan (a) to transform 2D image features into 1D sequential data.
    Based on Mamba (b), we design bidirectional Mamba (c) and prefix bidirectional Mamba module (d). Other details are shown in the right of this picture, includes flip, conditional flip (C-Flip), and two-norm (T-Norm).}
    \label{fig:architecture}
\end{figure}

As illustrated in Fig.~\ref{fig:overall_pipeline} (b), the inputs to the VTM-VAE decoder include image features, proprioception $\mathbf{h}_{t-p+1:t}$, and the latent vector $\mathbf{z}$.
The image features are extracted using ResNet-18 \cite{he2016deep}, which was pre-trained on ImageNet with BatchNorm layers frozen. The extracted image features have a dimension of $8 \times 10 \times 512$, which are then flattened into $80 \times 512$ using the proposed central-aware scan module, described later.
Subsequently, the image features, proprioception, and latent vector $\mathbf{z}$ are each projected to the same dimension $\mathbf{d}_{\text{model}}$, concatenated, and added with positional encoding. The resulting sequence is then fed into the proposed Bidirectional Mamba module, repeated $N_1$ times (Fig.~\ref{fig:architecture} (c)).
The representations are then concatenated with learnable parameters which correspond to the predicted actions. 
These learnable parameters share the same implementation with $\operatorname{CLS}$ tokens: end-to-end learned parameters that aggregate and propagate global information.
After passing through the proposed Prefix Bidirectional Mamba module, repeated $N_2$ times, (Fig.~\ref{fig:architecture} (d)). The learnable parameters are projected into $k$ actions, which are used for the reconstruction loss Eq. \ref{eq: loss}.

\subsubsection{Testing/Inference} \label{sec: real life inference}
During the inference stage (Fig.~\ref{fig:overall_pipeline} (c)), only the decoder is reserved, which takes the latent variable $\mathbf{z}$ sampled directly from a standard isotropic Gaussian distribution. Meanwhile, temporal ensemble \cite{zhao2023learning} is applied to better smooth the generated actions.

\subsection{Central-aware Mamba}
In this study, Central-aware Mamba collectively refers to Central-aware Scan, Bidirectional Mamba, and Prefix Bidirectional Mamba. Specifically, the Central-aware Scan converts 2D image features into 1D sequential features, while the Bidirectional Mamba and Prefix Bidirectional Mamba serve as the backbone of VTM-VAE.

Mamba \cite{gu2023mamba} is designed to process 1D signals using the Structured State Space Sequence Model (S4), achieving high-quality results with a limited number of parameters and optimized runtime efficiency. The architecture is illustrated in Fig.~\ref{fig:architecture} (b), in which the SSM follows the equations:
\begin{align}
    u(t) &= \overline{\mathbf{A}} u(t-1) + \overline{\mathbf{B}} x(t), \\
    y(t) &= \overline{\mathbf{C}} u(t),
\end{align}
where $\overline{\mathbf{A}} = \exp(\Delta \mathbf{A})$, $\overline{\mathbf{B}} = (\Delta \mathbf{A})^{-1} \left[ \exp(\Delta \mathbf{A}) - \mathbf{I} \right] \cdot (\Delta \mathbf{B})$, and $\overline{\mathbf{C}} = \mathbf{C}$. $u$, $x$, and $y$ are hidden state, input, and output, respectively.
$\mathbf{A}$ is a learnable parameter.
To incorporate an attention-like mechanism \cite{vaswani2017attention}, the parameters $\mathbf{B}, \mathbf{C},$ and $\Delta$ are input-dependent, obtained through a projection from the input data.

\subsubsection{Central-aware Scan}  
Since Mamba is designed to process 1D data, we propose the central-aware scan method to transform 2D images into 1D sequential data. In our task, the object of interest tends to be closer to the image center during grasping or releasing phases. Leveraging this characteristic, our central-aware scan progressively scans from the outer circle toward the inner circle, as shown in Fig.~\ref{fig:architecture} (a).

\subsubsection{Bidirectional Mamba \& Prefix Bidirectional Mamba}  
Mamba, as a uni-directional sequence modelling method, cannot propagate information bidirectionally-for example, $y_{1}$ cannot incorporate information from $y_{2}$. This limitation restricts its ability to comprehensively learn patterns from complex data types, such as images and videos.

To address this, we propose Bidirectional Mamba and Prefix Bidirectional Mamba. The input first undergoes layer normalization \cite{ba2016layer}, or in some cases, two separate layer normalizations. 
In the legend of Fig.~\ref{fig:architecture}, the two-layer normalization normalizes the input features (coloured bar) and the learnable parameters for predicted actions (gray bar) separately, as they follow different distributions. 
Next, the sequence passes through the Mamba layer with a residual connection. A flip or conditional flip (cond flip) is applied-where ``cond flip'' refers to flipping only the input features while keeping the rest action features unchanged, as shown in the legend 
of Fig.~\ref{fig:architecture}.

\section{Experiment}
\begin{figure}[t!]
    \centering  \includegraphics[width=0.485\textwidth]{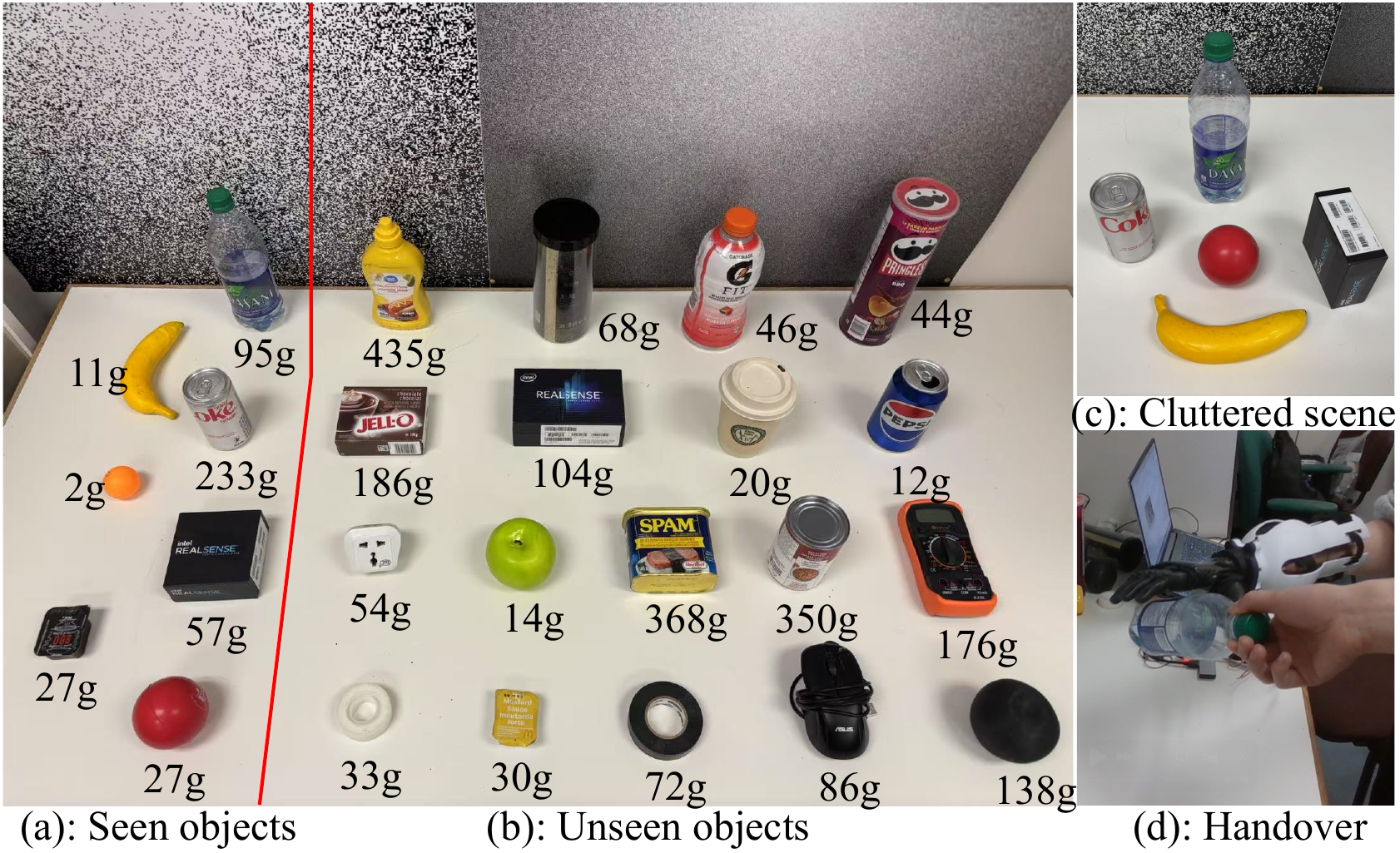}
    \caption{(a) Seven objects used for model training (Seen objects). (b) Eighteen unseen objects were used for evaluating the model's generalizability. (c) Cluttered scene used in this study, grasp and release the red ball at the same place. (d) Human-to-prosthetic handover task. The training data collector passes the bottle to the testing participant. 
    } 
    \label{fig:all_objects} 
\end{figure}

\subsection{Objects}

Figure~\ref{fig:all_objects} shows the objects and scenes used for both training data collection and testing. We chose a variety of everyday items with diverse shapes, textures, colours, and weights \cite{calli2015ycb}. Specifically, seven objects (Fig. \ref{fig:all_objects} (a)) were used for training data collection, while an additional 18 objects (Fig. \ref{fig:all_objects} (b)) were used during testing. Our goal is to train a model on only a few objects collected by one participant and then evaluate its generalizability across a wide variety of participants and everyday items. Furthermore, we tested the model in a cluttered scene (Fig. \ref{fig:all_objects} (c)) as well as in a handover scenario (Fig. \ref{fig:all_objects} (d)).

Figure \ref{fig:all_objects} shows the weight of each object.
In the seen objects, the weight ranges from 2 to 233 grams.
Notably, the ping-pong weighs only 2 grams.
In the unseen objects, the weight ranges from 14 to 435 grams.
In particular, the mustard bottle, located at the top left corner in Fig.~\ref{fig:all_objects} (b), weighs about twice as much as the heaviest seen object.

\subsection{Participants}

\begin{figure}[t!]
    \centering 
    \includegraphics[width=0.485\textwidth]{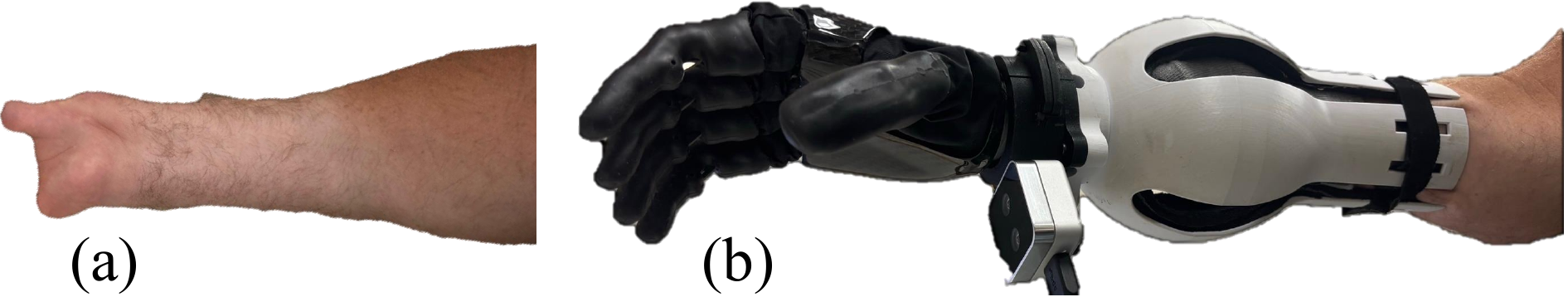}
    \caption{(a) The residual right forearm stump. (b) Participant wearing the prosthetic hand.}
    \label{fig:patient} 
\end{figure}

We recruited eleven healthy right-handed participants (nine males and two females) aged between 21 and 31 years (M = 26.1, SD = 3.2) with heights ranging from 153 cm to 186 cm (M = 173.9, SD = 10.6). 
In addition, we recruited a 61-year-old right-handed male with a limb difference (Fig.~\ref{fig:patient}), who is 175 cm tall.
The prosthetic hand control data was collected from one of the eleven healthy participants for model training, while the other participants took part in the testing experiments. 
Among them, one healthy testing participant was excluded due to an inability to follow instructions, attributed to insufficient upper arm strength.

Due to the large number of objects, the testing was divided into two sessions for healthy participants. In each session, objects were chosen at random, ensuring that all objects were covered collectively over the two sessions. The first session lasted 1-1.5 hours, and the second session lasted under one hour, the interval between sessions ranged from 5 to 12 days (M = 7.11, SD = 2.05).

Before beginning the experiments, participants watched an instructional video demonstrating the process of grasp, hold, and release to understand the prosthetic hand control procedure. They then completed several practice trials (typically fewer than ten, except for the ping-pong object) followed by ten consecutive recorded tests per object. This study was approved by the Memorial University Interdisciplinary Committee on Ethics in Human Research (20210316-SC), and all participants provided informed consent prior to the study.

\begin{figure}[t!]
    \centering 
    \includegraphics[width=0.485\textwidth]{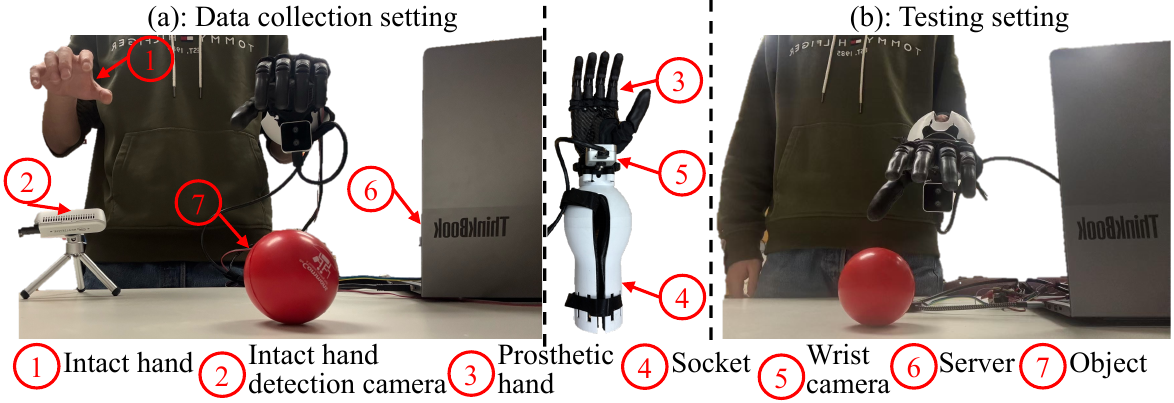}
    \caption{The training data collection and testing setting in this study, which simulate person with transradial limb loss.}
    \label{fig:teleoperation_setting} 
\end{figure}

\subsection{Training data collection setup and procedure}
\label{sec:teleoperation}

\subsubsection{Vision-based teleoperation system}
As shown in Fig.~\ref{fig:teleoperation_setting} (a), a prosthetic hand teleoperation system is designed to collect human demonstration data for imitation learning, where the left bare hand teleoperates the fingers of the prosthetic hand worn on the right hand to secure, hold or release the object, whereas the whole prosthetic hand movement is controlled by the right arm. 
The system runs at 30~Hz for intact hand detection \cite{lugaresi2019mediapipe}, wrist-view image recording and prosthetic hand control; and at approximately 50~Hz for recording proprioceptive information.

Specifically, the 3D-printed socket \textcircled{4} connects the prosthetic hand to the right intact hand and restricts the movement of the fingers of the right intact hand to mimic person with limb loss. Keypoints of the left intact hand \textcircled{1} are detected by a desktop camera \textcircled{2} and sent to a laptop \textcircled{6}, which calculates the fingertip position for configuring the prosthetic hand \textcircled{3} using a retargeting algorithm \cite{handa2020dexpilot}. The positions of the fingertip are then transmitted to the prosthetic hand \textcircled{3} via Universal Asynchronous Receiver/Transmitter (UART) and used as the target positions for the fingertips at the next timestep. Simultaneously, the vision data from the wrist camera \textcircled{5}, the proprioception data from the prosthetic hand \textcircled{3}, and the target fingertip positions for the prosthetic hand are recorded for imitation learning.

\subsubsection{Procedure}
During the entire data collection phase, the intact bare hand remained within the view of the desktop camera that detected the key points of the hand, with the wrist remaining relatively still; only the fingers move to configure the action that guided the prosthetic hand to grasp or release the object properly. The object was placed within a reachable range before each trial began. 
The fingertip forces data undergo zero-drift calibration, where the baseline reading was recorded when the hand was fully open.

In each trial, we followed several steps as outlined below:

\begin{enumerate} 
\item Rest State: The prosthetic hand is instructed to remain parallel to the desktop for two seconds. 
\item Reaching Phase: The participant moves their right arm to reach for the object. The hand should face down. 
\item Grasping Phase: Once the prosthetic hand is positioned and ready to grasp the object, the participant closes their left hand to teleoperate the prosthetic hand to grasp the object. 
\item Lifting Phase: After grasping the object, the participant lifts it to the height at which both the prosthetic hand and the participant’s forearm are parallel to the desktop. Simultaneously, the intact hand maintains a grasping pose to prevent the prosthetic hand from unintentionally opening and dropping the object. 
\item Holding Phase: The participant holds the object for approximately two seconds. During this time, the intact hand remains relatively still. 
\item Lowering Phase: Following the holding phase, the participant lowers the prosthetic hand toward the desktop while keeping the intact hand still. 
\item Releasing Phase: When the object is very close to the desktop, the participant opens the prosthetic hand with the guidance of the intact hand. 
\item Retrieving Phase: The participant retrieves the prosthetic hand back to the rest state, ensuring it remains parallel to the desktop. Simultaneously, the intact hand is kept open. \end{enumerate}

\subsection{Real-time testing setup and procedure}
\subsubsection{Test setting}
As shown in Fig.~\ref{fig:teleoperation_setting} (b), the desktop camera and the intact hand are no longer required, and the prosthetic hand is driven by the trained model with input from the wrist-view camera and proprioception.

\subsubsection{Procedure}

We use the same steps as described in the training data collection protocol, where the actions of the prosthetic hand are driven by the trained model instead of the intact hand. 
Each healthy participant collects 10 trials for each object, while 5 trials for the person with limb difference.

\subsection{Evaluation criteria}

We use four criteria in this study to quantify the trained model's success rate.
\begin{enumerate}
    \item SR$_{\text{r-g}}$: The ratio of successful reaching-grasping to total trials. Reaching-grasping is considered successful if Steps 2 and 3 are completed successfully and the subsequent lifting causes the object to move upward.

\item SR$_{\text{l-h-l}}$: The ratio of successful lifting-holding-lowering to successful reaching-grasping. Lifting-holding-lowering is deemed successful if Steps 4, 5 and 6 are completed successfully.

\item SR$_{\text{r-r}}$: The ratio of successful releasing-retrieving to successful lifting-holding-lowering. Releasing-retrieving is considered successful if Steps 7 and 8 are completed successfully.

\item SR$_{\text{trial}}$: The ratio of successful trials to total trials. A trial is considered successful if Steps 2 through 8 are completed successfully.
\end{enumerate}

\begin{table*}[htbp]
\centering
\caption{Success rate on seen objects, unseen objects. Each object is averaged among nine healthy participants and one person with limb difference.}
\centering
\begin{tabular}{|p{1.455cm}|*{8}{p{1.455cm}|}} 
\hline
\multicolumn{9}{|c|}{\textbf{Seen objects}}
\\ \hline
&
\includegraphics[scale=0.2]{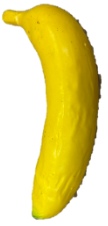} & 
\includegraphics[scale=0.25]{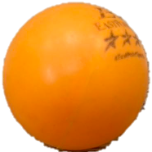} & 
\includegraphics[scale=0.2]{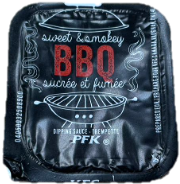} &
\includegraphics[scale=0.2]{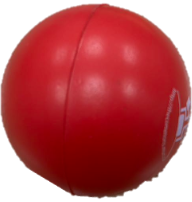} &
\includegraphics[scale=0.2]{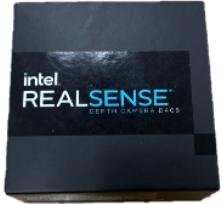} &
\includegraphics[scale=0.25]{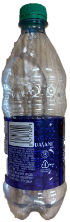} &
\includegraphics[scale=0.25]{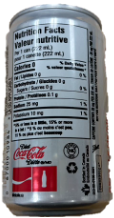} & 
\textbf{Average} 
\\ \hline
Object name & Banana & Ping-pong & BBQ sauce & Soft ball & D405 &  Bottle & Coke &     \\ \hline

SR$_{\text{r-g}}$  & $0.98_{\pm 0.04}$ & $0.98_{\pm 0.06}$ & $0.96_{\pm 0.07}$ & $0.99_{\pm 0.03}$ & $1.00_{\pm 0.00}$ & $1.00_{\pm 0.00}$ & $0.98_{\pm 0.04}$ & $0.98_{\pm 0.02}$  \\ \hline


SR$_{\text{l-h-l}}$ & $0.96_{\pm 0.07}$ & $0.84_{\pm 0.15}$ & $0.91_{\pm 0.10}$ & $0.98_{\pm 0.04}$ & $0.99_{\pm 0.03}$ & $0.99_{\pm 0.03}$ & $0.99_{\pm 0.03}$  & $0.95_{\pm 0.02}$  \\ \hline


SR$_{\text{r-r}}$ & $1.00_{\pm 0.00}$ & $1.00_{\pm 0.00}$ & $1.00_{\pm 0.00}$ & $0.99_{\pm 0.03}$ & $0.95_{\pm 0.07}$ & $1.00_{\pm 0.00}$ & $0.97_{\pm 0.06}$ & $0.99_{\pm 0.02}$ \\ \hline


SR$_{\text{trial}}$ & $0.94_{\pm 0.09}$ & $0.82_{\pm 0.15}$ & $0.87_{\pm 0.10}$ & $0.96_{\pm 0.06}$ & $0.94_{\pm 0.08}$ & $0.99_{\pm 0.03}$ & $0.94_{\pm 0.07}$ &  $0.92_{\pm 0.03}$ \\ \hline

\end{tabular}

\vspace{0.01cm}

\begin{tabular}{|c|*{9}{c|}} 

\hline
\multicolumn{10}{|c|}{\textbf{Unseen objects}}
\\ \hline
  &
\includegraphics[scale=0.045]{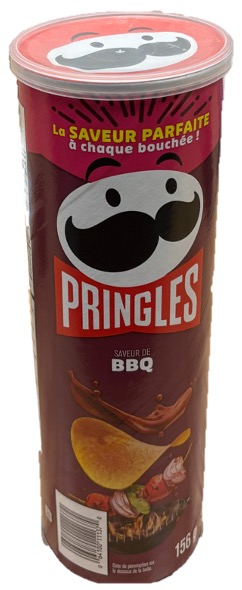} &
\includegraphics[scale=0.1]{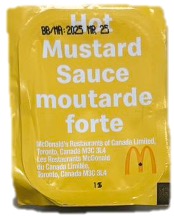} &
\includegraphics[scale=0.04]{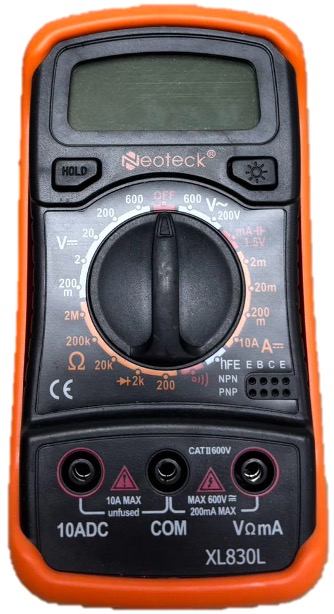} &
\includegraphics[scale=0.05]{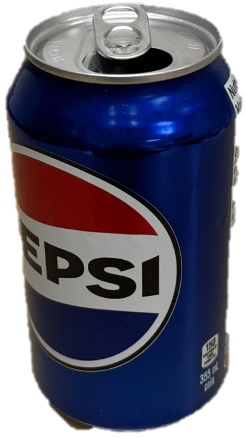} &
\includegraphics[scale=0.05]{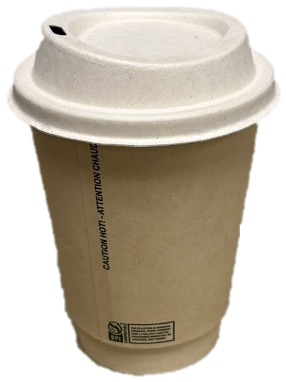} &
\includegraphics[scale=0.05]{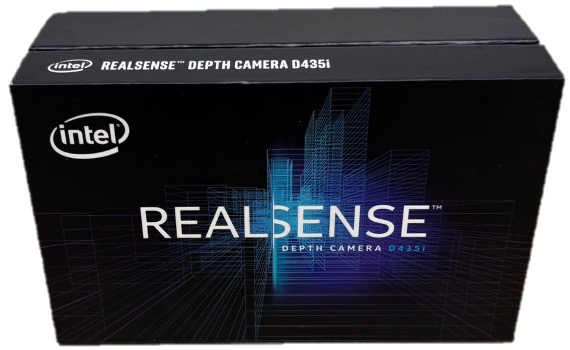} &
\includegraphics[scale=0.05]{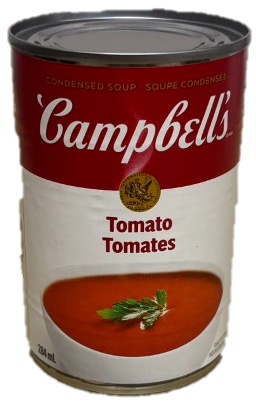} &
\includegraphics[scale=0.03]{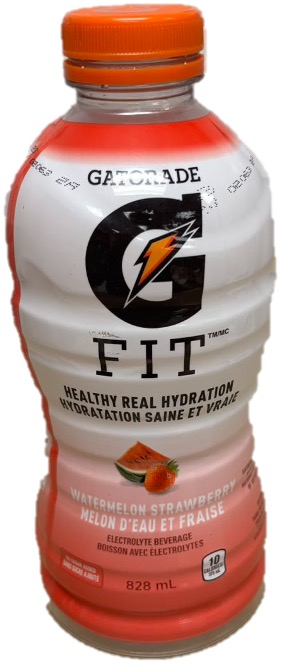} & 
\includegraphics[scale=0.05]{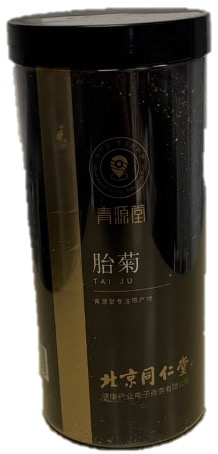} 
\\ \hline
Object name & Chips & Mustard-A & Multimeter & Pepsi & Paper cup & D435i box & Tomato can & Gatorade & Container \\ \hline

SR$_{\text{r-g}}$ & $1.00_{\pm 0.00}$ & $1.00_{\pm 0.00}$ & $0.98_{\pm 0.04}$ & $1.00_{\pm 0.00}$ & $0.99_{\pm 0.03}$ & $1.00_{\pm 0.00}$ & $0.95_{\pm 0.10}$ & $0.98_{\pm 0.06}$ & $0.98_{\pm 0.04}$ \\ \hline


SR$_{\text{l-h-l}}$ & $0.98_{\pm 0.06}$ & $0.82_{\pm 0.15}$ & $0.90_{\pm 0.23}$ & $0.95_{\pm 0.08}$ & $0.94_{\pm 0.07}$ & $0.83_{\pm 0.18}$ & $0.98_{\pm 0.06}$ & $0.99_{\pm 0.03}$ & $0.96_{\pm 0.07}$  \\ \hline


SR$_{\text{r-r}}$ & $0.95_{\pm 0.08}$ & $1.00_{\pm 0.00}$ & $0.99_{\pm 0.03}$ & $0.92_{\pm 0.15}$ & $0.93_{\pm 0.18}$ & $0.97_{\pm 0.07}$ & $0.95_{\pm 0.08}$ & $0.91_{\pm 0.08}$ & $0.97_{\pm 0.07}$  \\ \hline


SR$_{\text{trial}}$ & $0.93_{\pm 0.11}$ & $0.82_{\pm 0.15}$ & $0.88_{\pm 0.23}$ & $0.88_{\pm 0.18}$ & $0.86_{\pm 0.17}$ & $0.80_{\pm 0.17}$ & $0.88_{\pm 0.12}$ & $0.89_{\pm 0.12}$ & $0.91_{\pm 0.09}$ \\ \hline

\includegraphics[scale=0.08]{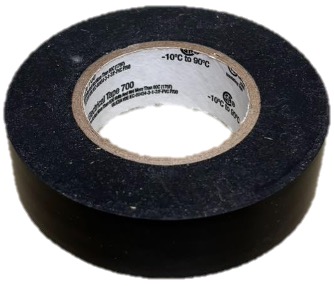} &
\includegraphics[scale=0.08]{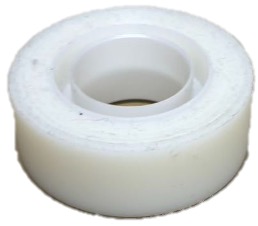} &
\includegraphics[scale=0.08]{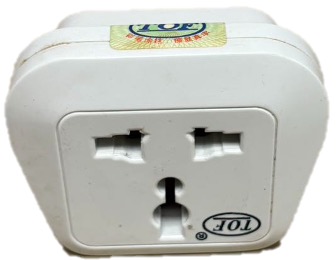} &
\includegraphics[scale=0.08]{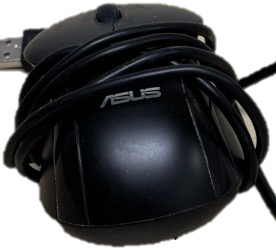} &
\includegraphics[scale=0.03]{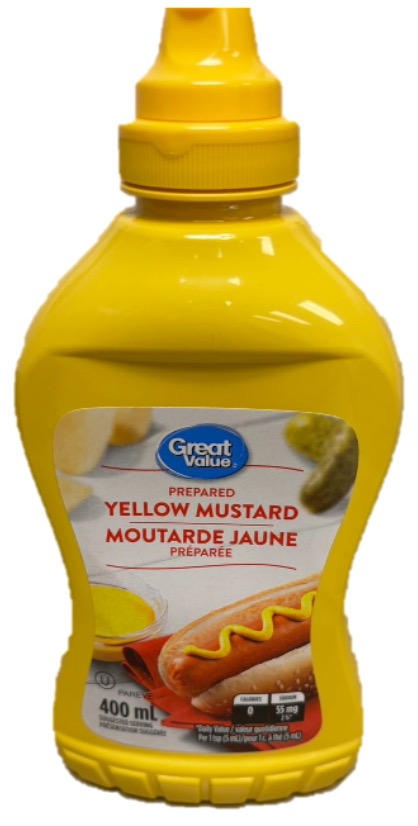} &
\includegraphics[scale=0.04]{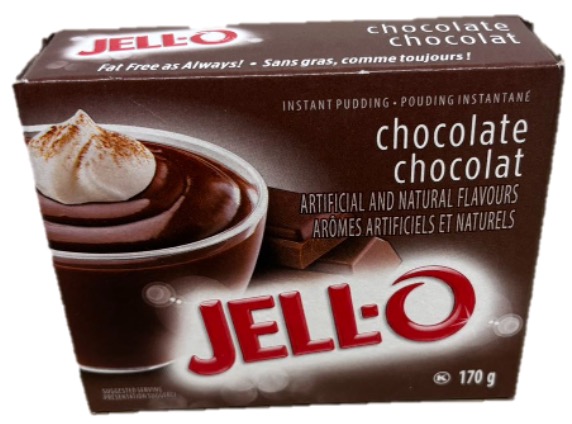} & 
\includegraphics[scale=0.04]{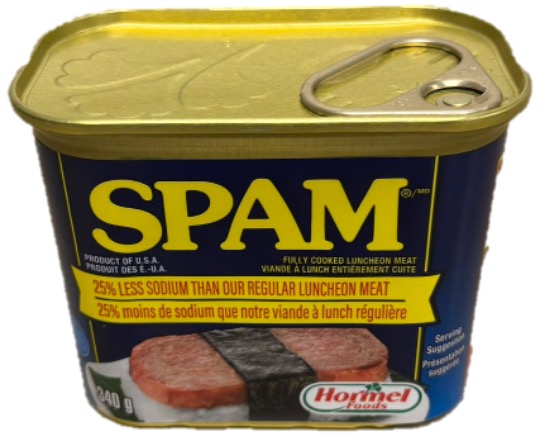} & 
\includegraphics[scale=0.1]{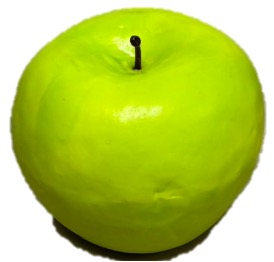} &
\includegraphics[scale=0.08] {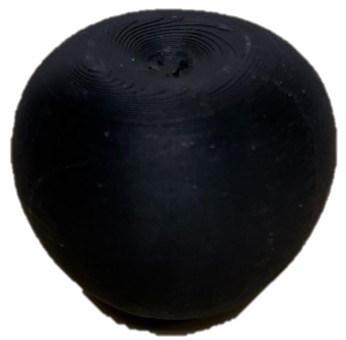} &
\textbf{Average}
\\ \hline
Tape-A  & Tape-B & Adapter & Mouse & Mustard-B &  Jello  & Spam & Apple-A & Apple-B &  \\ \hline

$0.99_{\pm 0.03}$ & $0.97_{\pm 0.06}$ & $1.00_{\pm 0.00}$ & $0.99_{\pm 0.03}$ & $0.98_{\pm 0.04}$ & $0.99_{\pm 0.03}$ & $0.98_{\pm 0.06}$ & $1.00_{\pm 0.00}$ & $1.00_{\pm 0.00}$ & $0.99_{\pm 0.01}$ \\ \hline


$0.99_{\pm 0.03}$ & $0.87_{\pm 0.15}$ & $0.95_{\pm 0.08}$ & $0.88_{\pm 0.17}$ & $0.91_{\pm 0.10}$ & $0.90_{\pm 0.16}$ & $0.97_{\pm 0.05}$ & $0.98_{\pm 0.04}$ & $0.97_{\pm 0.07}$ & $0.93_{\pm 0.02}$ \\ \hline


$1.00_{\pm 0.00}$ & $1.00_{\pm 0.00}$ & $0.99_{\pm 0.04}$ & $0.99_{\pm 0.04}$ & $1.00_{\pm 0.00}$ & $1.00_{\pm 0.00}$ & $0.90_{\pm 0.13}$ & $1.00_{\pm 0.00}$ & $1.00_{\pm 0.00}$ & $0.97_{\pm 0.02}$ \\ \hline


$0.98_{\pm 0.04}$ & $0.84_{\pm 0.14}$ & $0.94_{\pm 0.10}$ & $0.86_{\pm 0.17}$ & $0.89_{\pm 0.10}$ & $0.89_{\pm 0.16}$ & $0.85_{\pm 0.11}$ & $0.98_{\pm 0.04}$ & $0.97_{\pm 0.07}$ & $0.89_{\pm 0.03}$\\ \hline

\end{tabular}

\label{tab:real-world results}
\end{table*}

\begin{figure}[t!]
    \centering
    \includegraphics[width=0.485\textwidth]{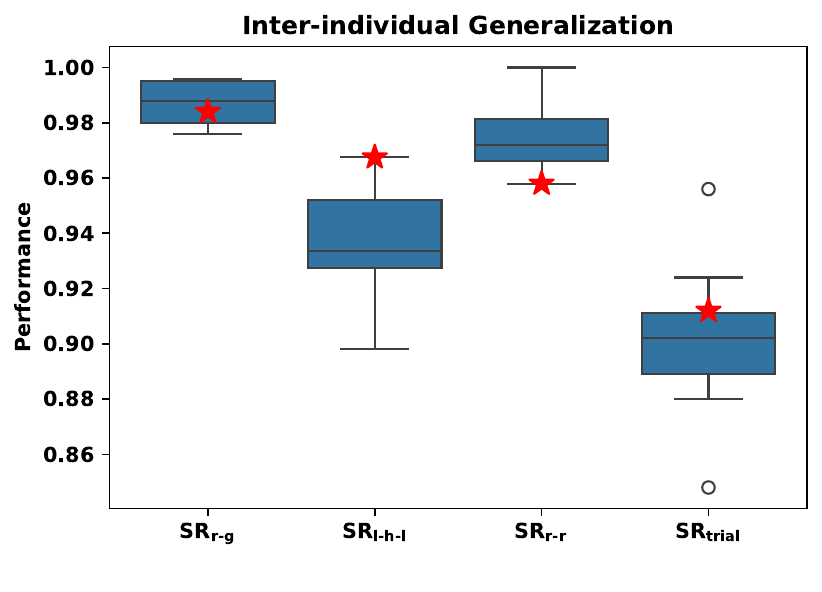}
    \caption{Average success rates of each criterion across nine healthy participants and one participant with limb difference. Each participant's data is an average of seen objects, and unseen objects. The red five-pointed star represents the person with limb difference.} \label{fig:cross participants} 
\end{figure}

\subsection{Model training details}

Based on small-scale experiments, we configured the model with hyperparameters $p=30$ in Eq. \ref{equ:encoder}, $k=20$ in Eq. \ref{equ:encoder}, $N_{1}=4$ in encoder, and $N_{2}=6$ in decoder; training was carried out for 50 epochs using the Adam optimizer with both a learning rate and weight decay of 0.0001; input images were resized to $320\times240$ pixels and processed in the batch of 128.

\section{Results}

For the model training, one single participant demonstrated 140 prosthetic hand control trials in total (20 trials per each training object for 7 objects) (Fig.~\ref{fig:all_objects} (a)) under the training data collection protocol.

For the model testing, a total of 2430 valid trials were collected from nine healthy participants (10 trials/object, 25 objects and extra two scenarios) (Fig.~\ref{fig:all_objects}).
Ten invalid trials were excluded, attributed to 5 hardware issues and 5 serious violations of the testing protocol leading to a total of 2420 trials data were included in the analysis.
For the person with limb difference, 125 trials were collected (5 trials/object, 25 objects).

\subsection{Inter-individual generalization}
Figure \ref{fig:cross participants} shows the resulting accuracies in terms of the four criteria across nine healthy participants and one individual with limb difference.
Each data point represents the performance of one participant across the seen objects, and unseen objects.

The mean and standard deviation for SR$_{\text{r-g}}$, SR$_{\text{l-h-l}}$, SR$_{\text{r-r}}$, and SR$_{\text{trial}}$ are $0.99_{\pm 0.01}$, $0.94_{\pm 0.02}$, $0.97_{\pm 0.01}$, $0.90_{\pm 0.03}$, respectively.
The high mean values for the four criteria showcase the impressive prosthetic hand control performance of our method, while the low standard deviation demonstrates its strong generalization across nine healthy individuals and one individual with limb difference. SR$_{\text{l-h-l}}$ is lower than SR$_{\text{r-g}}$ and SR$_{\text{r-r}}$, which indicates that the lifting-holding-lowering phase is harder to learn compared to the reaching-grasping and releasing-retrieving phases. SR$_{\text{trial}}$ is slightly low because it requires consecutive success from Step~2 to Step~8. Statistics of reasons for all failure trials are reported in Table \ref{tab:failure_reasons}.

\subsection{Inter-object generalization}
Table. \ref{tab:real-world results} shows quantitative results of seen objects and unseen objects, and each result is averaged by nine healthy participants one individual with limb difference.
The results indicate that our model is able to produce impressive actions under unseen objects and across participants.

\subsubsection{Seen objects}

On average, the algorithm achieved high success rates of $0.98_{\pm 0.02}$ for SR$_{\text{r-g}}$ and $0.99_{\pm 0.02}$ for SR$_{\text{r-r}}$, demonstrating the robustness of the action execution. However, SR$_{\text{l-h-l}}$ is slightly lower at $0.95_{\pm 0.02}$. 
To be specific, SR$_{\text{l-h-l}}$ is lower than SR$_{\text{r-g}}$ and SR$_{\text{r-r}}$ for 5 out of 7 objects (banana, ping-pong, BBQ sauce, softball, and bottle).
While the coke has higher SR$_{\text{l-h-l}}$  than SR$_{\text{r-g}}$ and SR$_{\text{r-r}}$.
Due to the small size of ping-pong and BBQ sauce, stable grasps are harder to achieve than other large-size objects.

\subsubsection{Unseen objects}
Our method has a strong ability to generalize to unseen objects.
On average, it has a similar SR$_{\text{r-g}}$ compared with seen objects.
However, SR$_{\text{l-h-l}}$ and SR$_{\text{r-r}}$ are slightly lower than the seen objects. 
The slightly lower SR$_{\text{l-h-l}}$ is mainly caused by mustard-a, multimeter, d435i box, tape-b, mouse, mustard-b, and jello.
The slightly lower SR$_{\text{r-r}}$ is mainly caused by chips, Pepsi, d435i box, tomato can, gatorade, and spam.


\subsection{Preliminary testing with cluttered scene and handover situation}
The cluttered scene in Fig.~\ref{fig:all_objects} (c) achieved $0.98_{\pm 0.06}$, $1.00_{\pm 0.00}$, $1.00_{\pm 0.00}$ and $0.98_{\pm 0.06}$ on four criteria.
The handover in Fig.~\ref{fig:all_objects} (d) achieved $1.00_{\pm 0.00}$ for four criteria.
These preliminary results indicate that our method has the potential to generalize to more unseen scenarios without any finetuning.

\section{Discussion}

This paper presents a novel method for prosthetic hand control that autonomously drives the prosthetic hand to grasp and release objects with an appropriate grip force using only a wrist-mounted camera. This approach significantly reduces the complexity of prosthetic hand control, 
leading to more intuitive operations. 

Our model demonstrates strong inter-individual generalization, as all results were obtained from the model trained with a single participant. This suggests that the model effectively captures transferable grasping strategies, highlighting its potential for real-world applications and greatly benefiting end-users by eliminating the need for individual-specific model training. 
A logical next step for this research would be to explore whether including more training data from multiple participants could further improve the model's performance and adaptability. 

Our model successfully learns grip configurations, timing of actions, and grip force application from demonstrations, enabling fully autonomous prosthetic hand control with impressive performance. However, to achieve more robust and reliable grasping, further challenges remain. 
Future research should explore how the model handles variations in object slipperiness, texture, and background conditions. 

Additionally, refining human intervention strategies could improve adaptability, though this aspect may not be essential for fully autonomous control. Addressing these aspects will be crucial for advancing prosthetic hand technology toward more natural and versatile grasping capabilities.

\begin{figure*}[h!]
    \centering
    \includegraphics[width=1.0\textwidth]{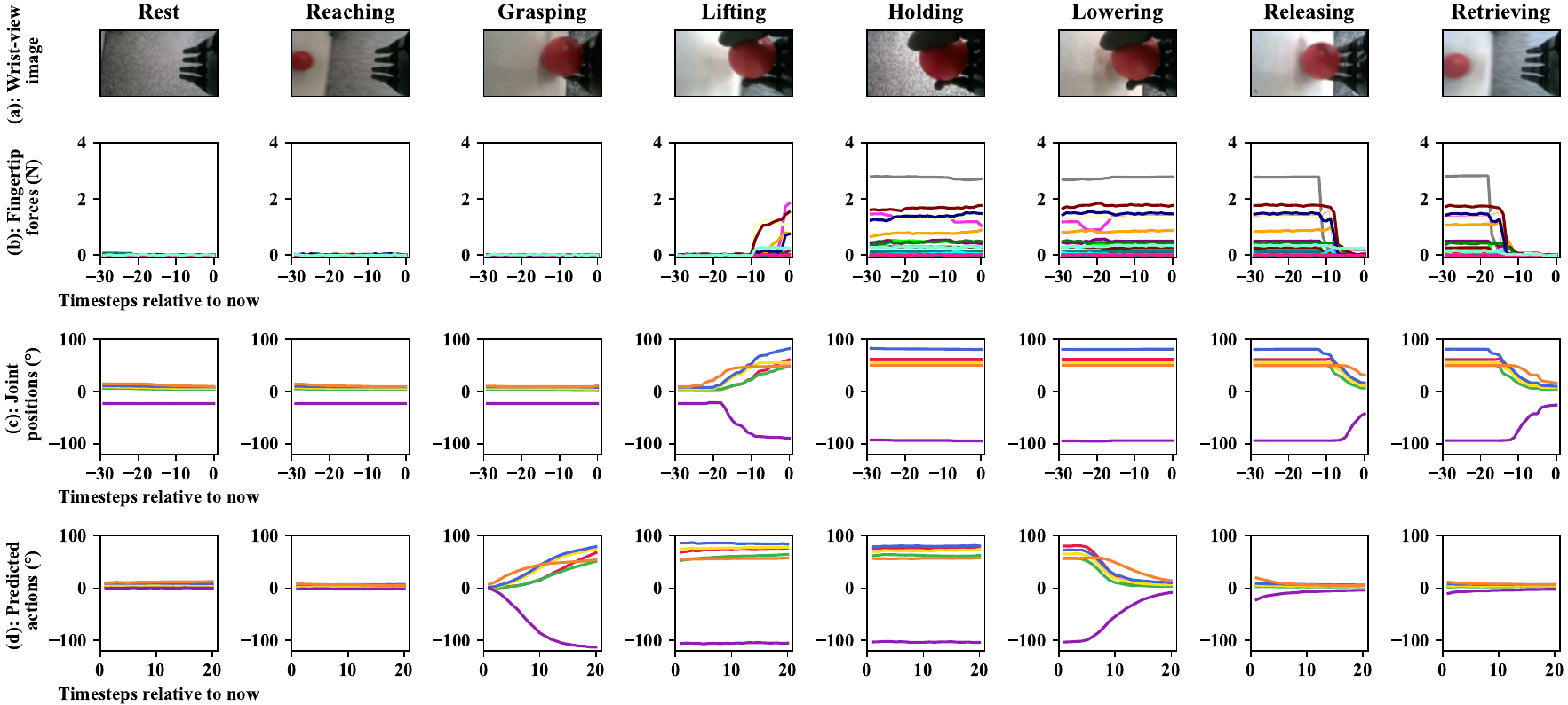}
    \caption{Inference visualization across eight phases: Each column represents a representative example moment during the phase, displaying a) the wrist-view image of the moment, b) historical fingertip forces (30 timesteps before the moment), c) historical joint positions (30 timesteps before the moment), and d) predicted actions (20 timesteps after the moment). In each subplot, the x-axis is anchored at "0" for the current timestep of the moment. Here, the symbol N represents the unit of force, and $^\circ$ denotes the unit of the joint angles-even though joint positions are used in line with traditional practices in robotic learning. In row (b), each subplot displays 30 lines representing individual force-sensitive resistors, while in rows (c) and (d), each subplot shows 6 lines corresponding to individual joints.}
    \label{fig:real-world-illustrate}
\end{figure*}

\subsection{Visualization of testing/inference sequence}
Figure \ref{fig:real-world-illustrate} shows an example trial's input and output of the algorithm across all grasping-lifting-releasing phases.
For simplicity, the current timestep is set to 0 in each phase.
The inputs include a wrist-view image $\mathbf{I}_0$, fingertip forces $\mathbf{f}_{-29:0}$, and joint positions $\mathbf{q}_{-29:0}$, as detailed in Sec. \ref{sec:VTM-VAE}.
The output comprises predicted actions $\mathbf{a}_{1:20}$, where each $\mathbf{a}_{i}$ corresponds to the predicted action for the $i$-th future timestep \cite{zhao2023learning}.

In the rest and reaching phases, the historical fingertip forces are zero (Fig. \ref{fig:real-world-illustrate} (b)), and the joint positions represent a fully open hand state (Fig. \ref{fig:real-world-illustrate} (c)). The predicted actions indicate that the hand will remain open over the next twenty timesteps (Fig. \ref{fig:real-world-illustrate} (d)).
During the grasping phase, the historical fingertip forces and joint positions remain the same as in the reaching phase; however, the predicted actions differ, signaling that the hand is preparing to close, because there is an object available to grasp.
In the lifting phase, changes in historical proprioceptive data (Fig. \ref{fig:real-world-illustrate} (b, c)) demonstrate that the prosthetic hand gradually tightens its grip around the object. The predicted actions indicate a continuation of grasping (Fig. \ref{fig:real-world-illustrate} (d)).
Throughout the holding phase, both historical proprioceptive data (Fig. \ref{fig:real-world-illustrate} (b, c)) and predicted actions (Fig. \ref{fig:real-world-illustrate} (d)) remain stable, showing that the grip is maintained steadily.
In the lowering phase, the historical proprioceptive data (Fig. \ref{fig:real-world-illustrate} (b, c)) continues to show stability, while the predicted actions (Fig. \ref{fig:real-world-illustrate} (d)) suggest the hand will begin to open, anticipating object placement onto a table.
In the releasing phase, historical proprioceptive data (Fig. \ref{fig:real-world-illustrate} (b, c)) reveal that the hand gradually opens, and the predicted actions (Fig. \ref{fig:real-world-illustrate} (d)) confirm the intention to continue opening, completing the object release.
Finally, in the retrieval phase, even though the wrist-view image resembles that of the reaching phase, the distinct historical proprioceptive information (Fig. \ref{fig:real-world-illustrate} (b, c)) gives rise to open actions.

\begin{table}[t!]
\centering
\caption{A Comparison with sEMG-based methods. The results of intra-subject work are from \cite{shi2024semi} with top-down grasping. "Semi-Auto" means semi-autonomous. "M." means Mouchoux et al.}
\begin{tabular}{|c|*{5}{c|}} 
\hline
Testing &
\textbf{Objects} &
\includegraphics[scale=0.15]{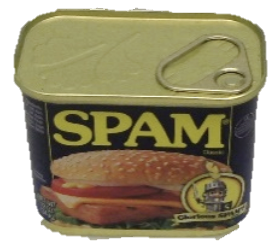} & 
\includegraphics[scale=0.15]{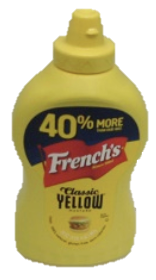} & 
\includegraphics[scale=0.3]{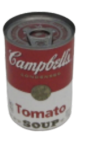} &
\includegraphics[scale=0.35]{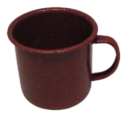} 
\\ \cline{1-6}
\multirow{3}{*}{Intra-subject} & EMG \cite{shi2024semi} & 66\% & 66\%  & 63\% & 46\% \\ \cline{2-6}
 & M. \cite{mouchoux2021artificial} & 97\% & 89\% & 90\% & 87\%\\ \cline{2-6}
&Semi-Auto \cite{shi2024semi} & 91\% & 89\%  & 94\% & 78\% \\ \hline
\multirow{2}{*}{Inter-subject} &
\textbf{Objects} &
\includegraphics[scale=0.05]{figure/objects/my_testing/Picture23.jpg} & 
\includegraphics[scale=0.025]{figure/objects/my_testing/Picture24.jpg} & 
\includegraphics[scale=0.05]{figure/objects/my_testing/Picture25.jpg} &
\includegraphics[scale=0.05]{figure/objects/my_testing/Picture9.jpg} 
\\ \cline{2-6}
& Ours & 98\% & 98\% & 95\% & 99\% \\ 
\hline
\end{tabular}
\label{tab:performance_semg}
\end{table}

\subsection{Compared with sEMG-based methods}

In Table \ref{tab:performance_semg}, we compare the results with sEMG-based methods, which need intra-subject models. 
The criteria for the compared three intra-subject methods are: a grasp is considered successful if the subject uses the correct preshape type to grasp the designated part of the object and maintains a slight shaking motion for 2 seconds without dropping it \cite{shi2024semi}.
As the compared methods only provide grasping success rates, we use our SR$_{\text{r-g}}$ to represent our grasping success rate.
Our results are comparable to those obtained by the alternative methods and, in some cases, even superior.
It is also worth mentioning that, our model has been tested with much more objects than these comparison methods.
Additionally, the sEMG-based methods use intra-subject testing, which needs to train a specific model for each participant.
Thus, our method has a competitive performance compared with the state-of-the-art semi-autonomous methods, and our method could be easier-to-use in real-world applications.

\newcommand{\xmark}{\ding{55}}
\newcommand{\cmark}{\ding{51}}
\begin{table}[t!]
    \centering
    \caption{A Descriptive comparison with other autonomous prosthetic hand control methods. "G" means grasp, "R" means release, "H" means human, and "O" means objects.}
    \begin{tabular}{|c|c|c|c|c|c|c|}
        \hline
        \multirow{2}{*}{Method} & \multicolumn{2}{c|}{Task} & \multicolumn{2}{c|}{Generalization} & \multirow{2}{*}{\shortstack{Real or \\ simulation}}  & \multirow{2}{*}{\shortstack{Camera \\ position}}\\
        \cline{2-5}
        & G & R & H & O & &  \\
        \hline
        Sharif et al. \cite{sharif2020towards} & \cmark & \xmark & \cmark & \xmark & Sim & \xmark \\ \hline
        Sharif et al. \cite{sharif2021end} & \cmark & \xmark & \cmark & \xmark & Real & Desk \\ \hline
        Xu et al. \cite{xu2025powered} & \cmark & \xmark & \cmark & \xmark & Real & Head \\ \hline
        Ours & \cmark & \cmark & \cmark & \cmark & Real & Wrist \\ \hline
    \end{tabular}
\label{tab:autonomous_comprasion}
\end{table}

\subsection{Compared with other autonomous grasping methods}

Table~\ref{tab:autonomous_comprasion} provides a comparison between the existing autonomous prosthetic hand grasping methods \cite{sharif2020towards}, \cite{sharif2021end}, \cite{xu2025powered} and our method. In real-world applications, automatic grasping and releasing are both essential. However, none of the three other studies support release tasks, which are not feasible for real-world applications. The method reported in \cite{sharif2020towards} only works in simulation. Sharif et al. \cite{sharif2021end} put a camera on the table, which limits the workspace. Xu et al. \cite{xu2025powered} put a camera on the head, and the user needs to look at the object. This method restricts the user's head movement. Our method contains grasp and release tasks that are closer to real-world applications, and the user need not fixate on the object.

\subsection{Failure cases analysis}

\begin{table}[h!]
\centering
\caption{Summary of Failed Cases. "A" means the failure mainly caused by the algorithm's miscalculation, and "U" means the failure mainly caused by the user's misoperation.}
\begin{tabular}{|c|c|c|}
\hline
Criteria & Failure description & Failure times \\ \hline
\multirow{6}{*}{SR$_{\text{r-g}}$}
&  Lifting before a secure grip. (U) &  5 \\ \cline{2-3}
&  Poor hand positioning (U) &  5 \\ \cline{2-3}
 & Closing too quickly (A)  &  3 \\ \cline{2-3}
& Object pushed away (A) &  6 \\ \cline{2-3}
 & Grasp opening instantly (A) &  6 \\ \cline{2-3}
& Insufficient hand closure (A) & 8 \\ \hline
\multirow{3}{*}{SR$_{\text{l-h-l}}$} & Grasp instability at step 4 (A)  & 71 \\ \cline{2-3}
&  Grasp instability at step 5 (A)& 45 \\ \cline{2-3}
&  Grasp instability at step 6 (A) & 36 \\ \hline
\multirow{2}{*}{SR$_{\text{r-r}}$} & Failure to open the hand (A) & 52 \\ \cline{2-3}
& Unintended grasp during retrieval (A) & 1  \\ \hline
\end{tabular}
\label{tab:failure_reasons}
\end{table}

We summarize all failure reasons of valid trials in Table \ref{tab:failure_reasons}.

\subsubsection{User-related failures}
\begin{enumerate}

\item Lifting before a secure grip:
The hand is lifted before achieving a firm grasp on the object, causing instability.
\item Poor hand positioning: 
    The prosthetic hand is improperly positioned, preventing it from successfully lifting the object after grasping.

\end{enumerate}

\subsubsection{Algorithm-related failures}
\begin{enumerate}
\item Closing too quickly:
The fingers close prematurely before the hand is properly positioned around the object, preventing a successful grasp.

\item Object pushed away: 
The applied force is unidirectional, causing the object to be unintentionally pushed away. 

\item Grasp opening instantly: 
After successfully grasping the object, the hand instantly opens.

\item Insufficient hand closure:
The fingers do not close tightly enough around the object, leading to an insecure grasp.

\item Grasp instability: 
The grasp is not secure, causing the object to slip and fall onto the table.

\item Failure to open the hand:
The algorithm fails to generate an open-hand action, preventing object release.

\item Unintended grasp during retrieval: 
When retracting the prosthetic hand after releasing the object, the hand unintentionally closes again, re-grasping the object.

\end{enumerate}

Our model was evaluated relatively strictly; for example, a grasp was classified as a failure if the hand assumed an incorrect posture during its initial approach to the object. In this case, the posture could be corrected by simply retrieving the hand a little bit back.
Actually, if the system was permitted to retract the hand slightly and attempt a second approach, the grasping accuracy could potentially reach 100\% for all trials. This highlights the potential for improved performance with minor adjustments to the evaluation protocol.

\subsection{How the model learns handling object weights variation}
As described in the experimental setup, the weights of the seven training objects range from 2 grams to 233 grams, while the testing objects range from 14 grams to 435 grams. The model effectively handles this range of weight variations by applying appropriate grip forces to securely grasp the target objects, indicating it has learned how to adjust grip force based on object weight. A possible explanation for this behavior is related to the teleoperation training method. During teleoperation, the demonstrator's actions (finger closures) typically continue until the fingertips fully contact each other, especially when handling heavier objects. Consequently, the corresponding grip forces, along with object-specific features, are implicitly captured and learned by the model.

\subsection{Expert demonstration collection in practice}

One question is how we can know whether the participant used the appropriate amount of force when picking up the object during the expert demonstration collection phase.

In this study, the participant underwent extensive practice in teleoperated grasping to ensure they applied sufficient force to prevent objects (i.e. Coke) from slipping, while avoiding excessive force that could damage or crush them (i.e. BBQ sauce). Trials in which the participant failed to grasp objects reliably-such as attempts with a banana or ping-pong ball-were excluded from the training dataset. In future work, we plan to integrate force feedback on the human bare hand so that demonstrators can receive real-time haptic cues.


\subsection{How to achieve more vision generalization ability}

Although this study eliminates sEMG, it relies on vision to infer control commands-introducing its own vision generalization challenges. Potential challenges are: 1) diverse approaching direction to objects; 2) diverse position and orientation of objects; 3) diverse height of users; 4) diverse objects and backgrounds. All of these challenges may result in different wrist-view images.

Fortunately, vision generalization problem has been extensively studied \cite{radford2021learning, hurst2024gpt, zhang2025robustdexgrasp} by collecting more data and training more powerful model.
To improve generalization ability, straightforward methods are gathering more data and employing more advanced machine-learning models.

\subsection{Challenging objects and possible solutions}

We list several explanations for why some objects pose challenges during grasping.
First, when grasping low-profile objects (i.e. a small-sized eraser), a prosthetic hand’s fingers tend to touch the table, making it difficult for users to find an optimal grasping position that ensures sufficient finger-object contact without unwanted table contact.
Second, small spherical objects like ping-pong balls demand precise equatorial alignment of the fingers before grasping with the prosthetic hand we used.
Other factors may originate from smoothness and softness of the objects.

In real life, these challenges could be partially handled using some grasping strategy. For example, when picking up a playing card from a table, one strategy is to slide the card to the table’s edge before grasping it. Moreover, human often use a pre-shape---adjusting the hand’s aperture and wrist orientation in advance---to facilitate a stable grasp \cite{jeannerod1984timing}.



\subsection{Beyond grasping-holding-releasing}

In practice, the demands on a prosthetic hand extend well beyond simple grasping: including pressing buttons, turning knobs, tying shoelaces, and countless other nuances. By moving our model beyond basic pick-and-place motions, we need to collect more data and propose more advanced algorithms.


\subsection{Practical, clinical and ethical implications}

We recognize that placing the large camera on the underside of the wrist can interfere with daily activities-human body may collide with it when resting the arm or making natural movements. To address this, future work will explore alternative mounting locations (e.g., the dorsal wrist or forearm) and use more compact, low-profile camera modules. These design changes aim to minimize bulk, reduce accidental impacts, and improve overall wearability in real-world settings.

All image processing will be performed locally, ensuring raw video never leaves the device. A hardware ``camera-ON" LED will clearly signal data capture.

Recognizing controllability of the prosthetic hand through the user, our future work will include a direct-control override that lets users switch instantly to manual operation. We will collaborate with clinical ethicists to draft user information sheets outlining system limitations and failure modes, and engage early with regulatory bodies (e.g., FDA, EMA) to chart a path toward clinical clearance.

\section{Conclusion, Limitations, and future work}

In this study, we demonstrate an intuitive and easy-to-use prosthetic hand system that operates autonomously, relying solely on wrist-mounted vision and the prosthetic hand's proprioception, without requiring any biosignals from the user. 
This biosignals-free property could reduce the operational complexity and burden associated with real-world prosthetic hand usage. 

However, the limitations still exist.
Firstly, this method has only been tested in a person with transradial limb loss setting, where the user can utilize their residual arm to move the hand for pick-up and placement tasks. Future work could expand to include a wider range of limb loss conditions. Secondly, the current testing has been limited to pick-up and place tasks involving only one hand. Future efforts will explore more complex manipulation tasks, including bimanual manipulation and a broader variety of functional grasps.
Thirdly, we will extend our approach to include a broader range of objects and backgrounds. This extension may involve leveraging simulation environments capable of generating an infinite variety of arrangements and combinations of diverse objects and scenarios.

\section{ACKNOWLEDGMENT}
The authors would like to thank Ann Shigeishi and the Dr. Leonard A. Miller Center team for invaluable assistance with amputee participant recruitment and prosthetic technical support.

\bibliography{main_first_revision_reference}

\begin{thebibliography}{10}
\providecommand{\url}[1]{#1}
\csname url@samestyle\endcsname
\providecommand{\newblock}{\relax}
\providecommand{\bibinfo}[2]{#2}
\providecommand{\BIBentrySTDinterwordspacing}{\spaceskip=0pt\relax}
\providecommand{\BIBentryALTinterwordstretchfactor}{4}
\providecommand{\BIBentryALTinterwordspacing}{\spaceskip=\fontdimen2\font plus
\BIBentryALTinterwordstretchfactor\fontdimen3\font minus \fontdimen4\font\relax}
\providecommand{\BIBforeignlanguage}[2]{{%
\expandafter\ifx\csname l@#1\endcsname\relax
\typeout{** WARNING: IEEEtran.bst: No hyphenation pattern has been}%
\typeout{** loaded for the language `#1'. Using the pattern for}%
\typeout{** the default language instead.}%
\else
\language=\csname l@#1\endcsname
\fi
#2}}
\providecommand{\BIBdecl}{\relax}
\BIBdecl

\bibitem{roche2014prosthetic}
A.~D. Roche, H.~Rehbaum, D.~Farina, and O.~C. Aszmann, ``Prosthetic myoelectric control strategies: a clinical perspective,'' \emph{Current Surgery Reports}, vol.~2, pp. 1--11, 2014.

\bibitem{nasr2021musclenet}
A.~Nasr, S.~Bell, J.~He, R.~L. Whittaker, N.~Jiang, C.~R. Dickerson, and J.~McPhee, ``Musclenet: mapping electromyography to kinematic and dynamic biomechanical variables by machine learning,'' \emph{Journal of Neural Engineering}, vol.~18, no.~4, p. 0460d3, 2021.

\bibitem{boostani2003evaluation}
R.~Boostani and M.~H. Moradi, ``Evaluation of the forearm emg signal features for the control of a prosthetic hand,'' \emph{Physiological Measurement}, vol.~24, no.~2, p. 309, 2003.

\bibitem{geethanjali2014low}
P.~Geethanjali and K.~Ray, ``A low-cost real-time research platform for emg pattern recognition-based prosthetic hand,'' \emph{IEEE/ASME Transactions on Mechatronics}, vol.~20, no.~4, pp. 1948--1955, 2014.

\bibitem{shi2024semi}
X.~Shi, W.~Guo, W.~Xu, Z.~Yang, and X.~Sheng, ``Semi-autonomous grasping control of prosthetic hand and wrist based on motion prior field,'' \emph{IEEE Robotics and Automation Letters}, 2024.

\bibitem{he2020vision}
Y.~He, R.~Kubozono, O.~Fukuda, N.~Yamaguchi, and H.~Okumura, ``Vision-based assistance for myoelectric hand control,'' \emph{IEEE Access}, vol.~8, pp. 201\,956--201\,965, 2020.

\bibitem{mcmullen2013demonstration}
D.~P. McMullen, G.~Hotson, K.~D. Katyal, B.~A. Wester, M.~S. Fifer, T.~G. McGee, A.~Harris, M.~S. Johannes, R.~J. Vogelstein, A.~D. Ravitz \emph{et~al.}, ``Demonstration of a semi-autonomous hybrid brain--machine interface using human intracranial eeg, eye tracking, and computer vision to control a robotic upper limb prosthetic,'' \emph{IEEE Transactions on Neural Systems and Rehabilitation Engineering}, vol.~22, no.~4, pp. 784--796, 2013.

\bibitem{ng2024development}
N.~Ng, ``Development of a semantic model and synthetic dataset for multi-grasp affordance detection for application to vision-based upper-limb prosthetic grasping,'' Master's thesis, University of Waterloo, 2024.

\bibitem{englehart2003robust}
K.~Englehart and B.~Hudgins, ``A robust, real-time control scheme for multifunction myoelectric control,'' \emph{IEEE Transactions on Biomedical Engineering}, vol.~50, no.~7, pp. 848--854, 2003.

\bibitem{hargrove2017myoelectric}
L.~J. Hargrove, L.~A. Miller, K.~Turner, and T.~A. Kuiken, ``Myoelectric pattern recognition outperforms direct control for transhumeral amputees with targeted muscle reinnervation: a randomized clinical trial,'' \emph{Scientific Reports}, vol.~7, no.~1, p. 13840, 2017.

\bibitem{simon2022user}
A.~M. Simon, K.~L. Turner, L.~A. Miller, B.~K. Potter, M.~D. Beachler, G.~A. Dumanian, L.~J. Hargrove, and T.~A. Kuiken, ``User performance with a transradial multi-articulating hand prosthesis during pattern recognition and direct control home use,'' \emph{IEEE Transactions on Neural Systems and Rehabilitation Engineering}, vol.~31, pp. 271--281, 2022.

\bibitem{boyer2023reducing}
M.~Boyer, L.~Bouyer, J.-S. Roy, and A.~Campeau-Lecours, ``Reducing noise, artifacts and interference in single-channel emg signals: A review,'' \emph{Sensors}, vol.~23, no.~6, p. 2927, 2023.

\bibitem{keenan2011coherence}
K.~G. Keenan, J.~D. Collins, W.~V. Massey, T.~J. Walters, and H.~D. Gruszka, ``Coherence between surface electromyograms is influenced by electrode placement in hand muscles,'' \emph{Journal of Neuroscience Methods}, vol. 195, no.~1, pp. 10--14, 2011.

\bibitem{vieira2023sensitivity}
T.~M. Vieira, G.~L. Cerone, A.~Botter, K.~Watanabe, and A.~D. Vigotsky, ``The sensitivity of bipolar electromyograms to muscle excitation scales with the inter-electrode distance.'' \emph{IEEE Transactions on Neural Systems and Rehabilitation Engineering}, 2023.

\bibitem{mogk2003crosstalk}
J.~P. Mogk and P.~J. Keir, ``Crosstalk in surface electromyography of the proximal forearm during gripping tasks,'' \emph{Journal of Electromyography and Kinesiology}, vol.~13, no.~1, pp. 63--71, 2003.

\bibitem{kong2010crosstalk}
Y.-K. Kong, M.~S. Hallbeck, and M.-C. Jung, ``Crosstalk effect on surface electromyogram of the forearm flexors during a static grip task,'' \emph{Journal of Electromyography and Kinesiology}, vol.~20, no.~6, pp. 1223--1229, 2010.

\bibitem{jarque2024does}
N.~J. Jarque-Bou, M.~Vergara, and J.~L. Sancho-Bru, ``Does exerting grasps involve a finite set of muscle patterns? a study of intra-and intersubject variability of forearm semg signals in seven grasp types,'' \emph{IEEE Transactions on Neural Systems and Rehabilitation Engineering}, 2024.

\bibitem{jiang2012myoelectric}
N.~Jiang, S.~Dosen, K.-R. Muller, and D.~Farina, ``Myoelectric control of artificial limbs—is there a need to change focus?[in the spotlight],'' \emph{IEEE Signal Processing Magazine}, vol.~29, no.~5, pp. 152--150, 2012.

\bibitem{wang2021effect}
J.~Wang, M.~Pang, P.~Yu, B.~Tang, K.~Xiang, and Z.~Ju, ``Effect of muscle fatigue on surface electromyography-based hand grasp force estimation,'' \emph{Applied Bionics and Biomechanics}, vol. 2021, no.~1, p. 8817480, 2021.

\bibitem{fang2022simultaneous}
B.~Fang, C.~Wang, F.~Sun, Z.~Chen, J.~Shan, H.~Liu, W.~Ding, and W.~Liang, ``Simultaneous semg recognition of gestures and force levels for interaction with prosthetic hand,'' \emph{IEEE Transactions on Neural Systems and Rehabilitation Engineering}, vol.~30, pp. 2426--2436, 2022.

\bibitem{dovsen2010cognitive}
S.~Do{\v{s}}en, C.~Cipriani, M.~Kosti{\'c}, M.~Controzzi, M.~C. Carrozza, and D.~B. Popovi{\'c}, ``Cognitive vision system for control of dexterous prosthetic hands: experimental evaluation,'' \emph{Journal of Neuroengineering and Rehabilitation}, vol.~7, pp. 1--14, 2010.

\bibitem{zandigohar2024multimodal}
M.~Zandigohar, M.~Han, M.~Sharif, S.~Y. G{\"u}nay, M.~P. Furmanek, M.~Yarossi, P.~Bonato, C.~Onal, T.~Pad{\i}r, D.~Erdo{\u{g}}mu{\c{s}} \emph{et~al.}, ``Multimodal fusion of emg and vision for human grasp intent inference in prosthetic hand control,'' \emph{Frontiers in Robotics and AI}, vol.~11, p. 1312554, 2024.

\bibitem{starke2022semi}
J.~Starke, P.~Weiner, M.~Crell, and T.~Asfour, ``Semi-autonomous control of prosthetic hands based on multimodal sensing, human grasp demonstration and user intention,'' \emph{Robotics and Autonomous Systems}, vol. 154, p. 104123, 2022.

\bibitem{castro2022continuous}
M.~N. Castro and S.~Dosen, ``Continuous semi-autonomous prosthesis control using a depth sensor on the hand,'' \emph{Frontiers in Neurorobotics}, vol.~16, p. 814973, 2022.

\bibitem{mouchoux2021artificial}
J.~Mouchoux, S.~Carisi, S.~Dosen, D.~Farina, A.~F. Schilling, and M.~Markovic, ``Artificial perception and semiautonomous control in myoelectric hand prostheses increases performance and decreases effort,'' \emph{IEEE Transactions on Robotics}, vol.~37, no.~4, pp. 1298--1312, 2021.

\bibitem{wang2022phase}
S.~Wang, J.~Zheng, B.~Zheng, and X.~Jiang, ``Phase-based grasp classification for prosthetic hand control using semg,'' \emph{Biosensors}, vol.~12, no.~2, p.~57, 2022.

\bibitem{wang2022integrating}
S.~Wang, J.~Zheng, Z.~Huang, X.~Zhang, V.~Prado~da Fonseca, B.~Zheng, and X.~Jiang, ``Integrating computer vision to prosthetic hand control with semg: Preliminary results in grasp classification,'' \emph{Frontiers in Robotics and AI}, vol.~9, p. 948238, 2022.

\bibitem{xu2025powered}
Y.~Xu, X.~Wang, J.~Li, X.~Zhang, F.~Li, Q.~Gao, C.~Fu, and Y.~Leng, ``A powered prosthetic hand with vision system for enhancing the anthropopathic grasp,'' \emph{IEEE Transactions on Neural Systems and Rehabilitation Engineering}, 2025.

\bibitem{sharif2020towards}
M.~Sharif, D.~Erdogmus, C.~Amato, and T.~Padir, ``Towards end-to-end control of a robot prosthetic hand via reinforcement learning,'' in \emph{2020 8th IEEE RAS/EMBS International Conference for Biomedical Robotics and Biomechatronics}.\hskip 1em plus 0.5em minus 0.4em\relax IEEE, 2020, pp. 641--647.

\bibitem{sharif2021end}
M.~Sharif, D.~Erdogmus, C.~Amato, and T.~Padir, ``End-to-end grasping policies for human-in-the-loop robots via deep reinforcement learning,'' in \emph{2021 IEEE International Conference on Robotics and Automation}.\hskip 1em plus 0.5em minus 0.4em\relax IEEE, 2021, pp. 2768--2774.

\bibitem{montagnani2015exploiting}
F.~Montagnani, M.~Controzzi, and C.~Cipriani, ``Exploiting arm posture synergies in activities of daily living to control the wrist rotation in upper limb prostheses: A feasibility study,'' in \emph{2015 37th Annual International Conference of the IEEE Engineering in Medicine and Biology Society}.\hskip 1em plus 0.5em minus 0.4em\relax IEEE, 2015, pp. 2462--2465.

\bibitem{kuhn2024synergy}
J.~K{\"u}hn, T.~Hu, A.~T{\"o}dtheide, E.~Pozo~Fortuni{\'c}, E.~Jensen, and S.~Haddadin, ``The synergy complement control approach for seamless limb-driven prostheses,'' \emph{Nature Machine Intelligence}, vol.~6, no.~4, pp. 481--492, 2024.

\bibitem{merad2020assessment}
M.~Merad, E.~De~Montalivet, M.~Legrand, E.~Mastinu, M.~Ortiz-Catalan, A.~Touillet, N.~Martinet, J.~Paysant, A.~Roby-Brami, and N.~Jarrasse, ``Assessment of an automatic prosthetic elbow control strategy using residual limb motion for transhumeral amputated individuals with socket or osseointegrated prostheses,'' \emph{IEEE Transactions on Medical Robotics and Bionics}, vol.~2, no.~1, pp. 38--49, 2020.

\bibitem{bennett2017imu}
D.~A. Bennett and M.~Goldfarb, ``Imu-based wrist rotation control of a transradial myoelectric prosthesis,'' \emph{IEEE Transactions on Neural Systems and Rehabilitation Engineering}, vol.~26, no.~2, pp. 419--427, 2017.

\bibitem{pilarski2012dynamic}
P.~M. Pilarski, M.~R. Dawson, T.~Degris, J.~P. Carey, and R.~S. Sutton, ``Dynamic switching and real-time machine learning for improved human control of assistive biomedical robots,'' in \emph{2012 4th IEEE RAS \& EMBS International Conference on Biomedical Robotics and Biomechatronics}.\hskip 1em plus 0.5em minus 0.4em\relax IEEE, 2012, pp. 296--302.

\bibitem{mastinu2024explorations}
E.~Mastinu, A.~Coletti, J.~van~den Berg, and C.~Cipriani, ``Explorations of autonomous prosthetic grasping via proximity vision and deep learning,'' \emph{IEEE Transactions on Medical Robotics and Bionics}, 2024.

\bibitem{heo2023proximity}
S.-H. Heo and H.-S. Park, ``Proximity perception-based grasping intelligence: Toward the seamless control of a dexterous prosthetic hand,'' \emph{IEEE/ASME Transactions on Mechatronics}, vol.~29, no.~3, pp. 2079--2090, 2023.

\bibitem{handa2020dexpilot}
A.~Handa, K.~Van~Wyk, W.~Yang, J.~Liang, Y.-W. Chao, Q.~Wan, S.~Birchfield, N.~Ratliff, and D.~Fox, ``Dexpilot: Vision-based teleoperation of dexterous robotic hand-arm system,'' in \emph{2020 IEEE International Conference on Robotics and Automation}.\hskip 1em plus 0.5em minus 0.4em\relax IEEE, 2020, pp. 9164--9170.

\bibitem{ke2020rethinking}
G.~Ke, D.~He, and T.-Y. Liu, ``Rethinking positional encoding in language pre-training,'' \emph{arXiv preprint arXiv:2006.15595}, 2020.

\bibitem{kingma2013auto}
D.~P. Kingma, M.~Welling \emph{et~al.}, ``Auto-encoding variational bayes,'' 2013.

\bibitem{he2016deep}
K.~He, X.~Zhang, S.~Ren, and J.~Sun, ``Deep residual learning for image recognition,'' in \emph{Proceedings of the IEEE Conference on Computer Vision and Pattern Recognition}, 2016, pp. 770--778.

\bibitem{zhao2023learning}
T.~Z. Zhao, V.~Kumar, S.~Levine, and C.~Finn, ``Learning fine-grained bimanual manipulation with low-cost hardware,'' \emph{arXiv preprint arXiv:2304.13705}, 2023.

\bibitem{gu2023mamba}
A.~Gu and T.~Dao, ``Mamba: Linear-time sequence modeling with selective state spaces,'' \emph{arXiv preprint arXiv:2312.00752}, 2023.

\bibitem{vaswani2017attention}
A.~Vaswani, N.~Shazeer, N.~Parmar, J.~Uszkoreit, L.~Jones, A.~N. Gomez, {\L}.~Kaiser, and I.~Polosukhin, ``Attention is all you need,'' \emph{Advances in Neural Information Processing Systems}, vol.~30, 2017.

\bibitem{ba2016layer}
J.~L. Ba, ``Layer normalization,'' \emph{arXiv preprint arXiv:1607.06450}, 2016.

\bibitem{calli2015ycb}
B.~Calli, A.~Singh, A.~Walsman, S.~Srinivasa, P.~Abbeel, and A.~M. Dollar, ``The ycb object and model set: Towards common benchmarks for manipulation research,'' in \emph{2015 International Conference on Advanced Robotics}.\hskip 1em plus 0.5em minus 0.4em\relax IEEE, 2015, pp. 510--517.

\bibitem{lugaresi2019mediapipe}
C.~Lugaresi, J.~Tang, H.~Nash, C.~McClanahan, E.~Uboweja, M.~Hays, F.~Zhang, C.-L. Chang, M.~G. Yong, J.~Lee \emph{et~al.}, ``Mediapipe: A framework for building perception pipelines,'' \emph{arXiv preprint arXiv:1906.08172}, 2019.

\bibitem{radford2021learning}
A.~Radford, J.~W. Kim, C.~Hallacy, A.~Ramesh, G.~Goh, S.~Agarwal, G.~Sastry, A.~Askell, P.~Mishkin, J.~Clark \emph{et~al.}, ``Learning transferable visual models from natural language supervision,'' in \emph{International Conference on Machine Learning}.\hskip 1em plus 0.5em minus 0.4em\relax PmLR, 2021, pp. 8748--8763.

\bibitem{hurst2024gpt}
A.~Hurst, A.~Lerer, A.~P. Goucher, A.~Perelman, A.~Ramesh, A.~Clark, A.~Ostrow, A.~Welihinda, A.~Hayes, A.~Radford \emph{et~al.}, ``Gpt-4o system card,'' \emph{arXiv preprint arXiv:2410.21276}, 2024.

\bibitem{zhang2025robustdexgrasp}
H.~Zhang, Z.~Wu, L.~Huang, S.~Christen, and J.~Song, ``Robustdexgrasp: Robust dexterous grasping of general objects,'' \emph{arXiv preprint arXiv:2504.05287}, 2025.

\bibitem{jeannerod1984timing}
M.~Jeannerod, ``The timing of natural prehension movements,'' \emph{Journal of Motor Behavior}, vol.~16, no.~3, pp. 235--254, 1984.

\end{thebibliography}
\bibliographystyle{IEEEtran}
\end{document}